\def\year{2022}\relax
\documentclass[letterpaper]{article} 
\usepackage{aaai22}  
\usepackage{times}  
\usepackage{helvet}  
\usepackage{courier}  
\usepackage[hyphens]{url}  
\usepackage{graphicx} 
\usepackage{natbib}  
\usepackage{caption} 
\DeclareCaptionStyle{ruled}{labelfont=normalfont,labelsep=colon,strut=off} 
\usepackage{algorithm}
\usepackage{algorithmic}

\usepackage{epsfig}
\usepackage{amsmath}
\usepackage{amssymb}
\usepackage{changes}

\def\bv #1{\boldsymbol{\rm{#1}}}
\def\bm #1{\boldsymbol{#1}}
\usepackage{bbm}
\usepackage{multirow}
\usepackage{subfigure}
\usepackage{booktabs}

\newcommand{\Tref}[1]{Tab.~\ref{#1}}
\newcommand{\Eref}[1]{Equ.~(\ref{#1})}
\newcommand{\Fref}[1]{Fig.~\ref{#1}}

\newcommand{\tabincell}[2]{\begin{tabular}{@{}#1@{}}#2\end{tabular}}

\makeatletter
\DeclareRobustCommand\onedot{\futurelet\@let@token\@onedot}
\def\@onedot{.}
\def\eg{\textit{e.g}\onedot} 
\def\ie{\textit{i.e}\onedot}

\def\etal{\textit{et al}\onedot~}

\def\NA{\textit{N/A}}
\definecolor{sgreen}{rgb}{0.2,0.6,0.15}
\definecolor{sblue}{rgb}{0,0.3,0.9}
\makeatother

\usepackage{newfloat}
\usepackage{listings}
\floatstyle{ruled}
\newfloat{listing}{tb}{lst}{}
\floatname{listing}{Listing}
\title{PureGaze: Purifying Gaze Feature for Generalizable Gaze Estimation}
\author{
   	Yihua Cheng\textsuperscript{\rm 1}, Yiwei Bao\textsuperscript{\rm 1}, Feng Lu\textsuperscript{\rm 1, 2}\thanks{ Corresponding author.}
}
\affiliations{ \textsuperscript{\rm 1}State Key Laboratory of VR Technology and Systems, School of CSE, Beihang University\\
	\textsuperscript{\rm 2}Peng Cheng Laboratory, Shenzhen, China\\
\{yihua\_c, baoyiwei, lufeng\}@buaa.edu.cn
}

\usepackage{bibentry}

\begin{document}

\maketitle

\begin{abstract}
Gaze estimation methods learn eye gaze from facial features. 
However, among rich information in the facial image, real gaze-relevant features only correspond to subtle changes in eye region, while other gaze-irrelevant features like illumination, personal appearance and even facial expression may affect the learning in an unexpected way. 
This is a major reason why existing methods show significant performance degradation in cross-domain/dataset evaluation.
In this paper, we tackle the cross-domain problem in gaze estimation.
Different from common domain adaption methods, we propose a domain generalization method to improve the cross-domain performance without touching target samples.
The domain generalization is realized by gaze feature purification.
We eliminate gaze-irrelevant factors such as illumination and identity to improve the cross-domain performance.
We design a plug-and-play self-adversarial framework for the gaze feature purification.
The framework enhances not only our baseline but also existing gaze estimation methods directly and significantly.
To the best of our knowledge, we are the first to propose domain generalization methods in gaze estimation.
Our method achieves not only state-of-the-art performance among typical gaze estimation methods but also competitive results among domain adaption methods.
The code is released in \url{https://github.com/yihuacheng/PureGaze}.

\end{abstract}
	\section{Introduction}
Human gaze implicates important cues for understanding human cognition~\cite{Rahal_2019_ESP} and behavior~\cite{Dias_2020_WACV}.
It enables researchers to gain insights into many areas such as saliency detection~\cite{Wang_2019_TPAMI, Wang_2018_attentionpre}, virtual reality~\cite{Xu_2018_CVPR} and first-person video analysis~\cite{Yu_2020_TPAMI}.
Recently, appearance-based gaze estimation with deep learning becomes a hot topic.
They leverage convolutional neural networks (CNNs) to estimate gaze from human appearance~\cite{Park_2018_ECCV, Xiong_2019_CVPR}, and achieve accurate performance.
\begin{figure}[htp]
	\begin{center}
		\includegraphics[width=\columnwidth]{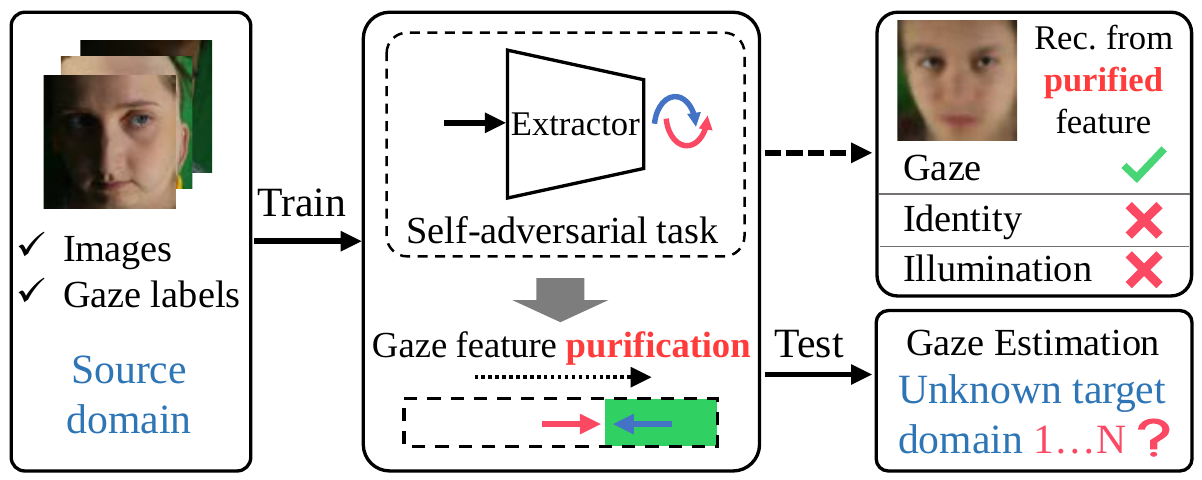}	
	\end{center}
	\caption{We propose a domain-generalization framework for gaze estimation. 
		Our method is only trained in the source domain and brings improvement in all unknown target domains.  
		The key idea of our method is to purify the gaze feature with self-adversarial framework.
		The visualization result shows gaze-irrelevant factors such as illumination and identity are eliminated from the extracted feature.}
	\label{fig:pipeline}
\end{figure}

\begin{figure*}[htp]
	\begin{center}
		\includegraphics[width=2\columnwidth]{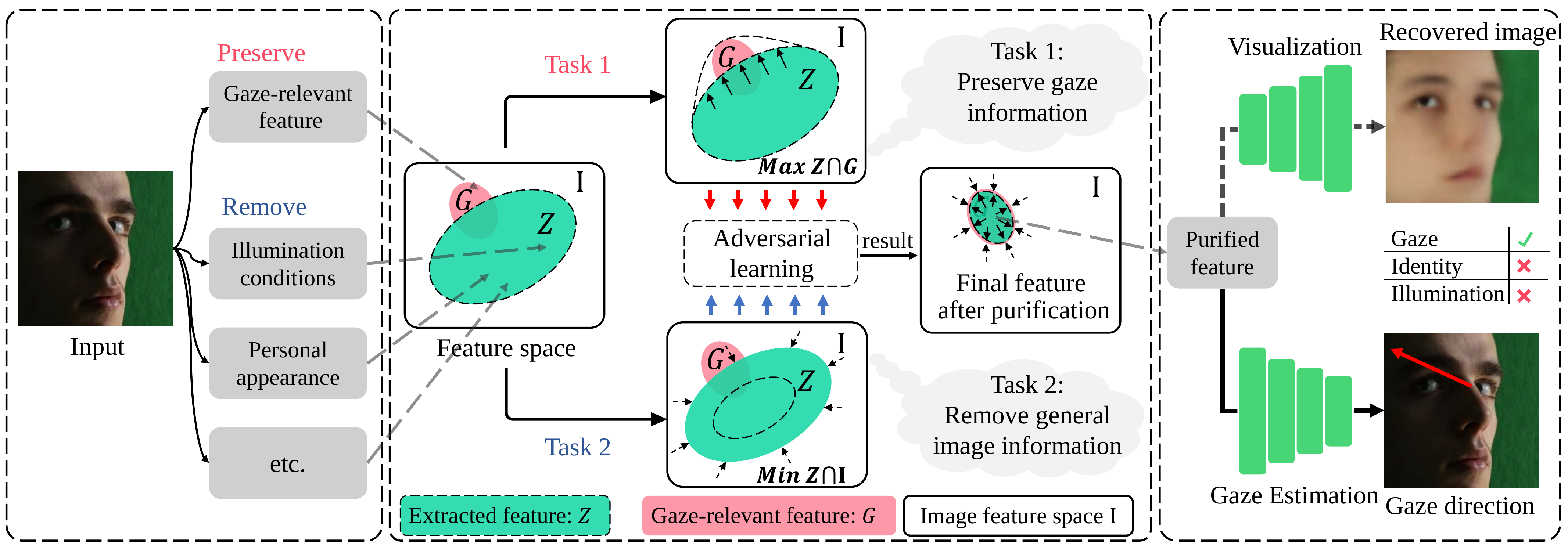}	
	\end{center}
	\caption{Overview of the gaze feature purification. Our goal is to preserve the gaze-relevant feature and eliminate gaze-irrelevant features. Therefore, we define two tasks, which are to preserve gaze information and to remove general facial image information. The two tasks are not cooperative but adversarial to purify feature.  
	Simultaneously optimizing the two tasks, we implicitly purify the gaze feature without defining gaze-irrelevant feature.}
	\label{fig:overview}
\end{figure*}

CNN-based gaze estimation requires a large number of samples for training. But collecting gaze sample is difficult and time-consuming.
This challenge can be ignored in a fixed environment, but becomes a bottleneck when gaze estimation is required in a new environment.
The changed environment brings many unexpected factors such as different illumination, thus degrades the performance of pre-trained model.
Recent methods usually handle the cross-environment problem\footnote{We refer it as cross-domain/dataset problem in the rest.} as a domain adaption problem.
Researches aim to adapt the model trained in source domains to target domains~\cite{Zhang_2018_CHI,Kellnhofer_2019_ICCV}.
However, these methods usually require target samples and time-consuming setup.
These requirements greatly harm the flexibility of methods.

In this paper, we innovate a new direction to solve the problem.
We propose a domain-generalization method for improving the cross-domain performance.
Our method does not require any images or labels in target domains, but aims to learn a generalized model in the source domain for any ``unseen" target domains.
We notice \textit{the intrinsic gaze pattern is similar in all domains, but there are domain differences in gaze-irrelevant factors such as illumination and identity.} 
These factors are usually domain-specific, and directly blend in captured images.
The in-depth fusion makes these factors difficult to be eliminated during feature extraction.
As a result, the trained model usually learns a joint distribution of gaze and these factors,~\ie, overfit in source domain, and naturally cannot perform well in target domains. 

\begin{figure*}[t]
	\begin{center}
		\includegraphics[width=1.95\columnwidth]{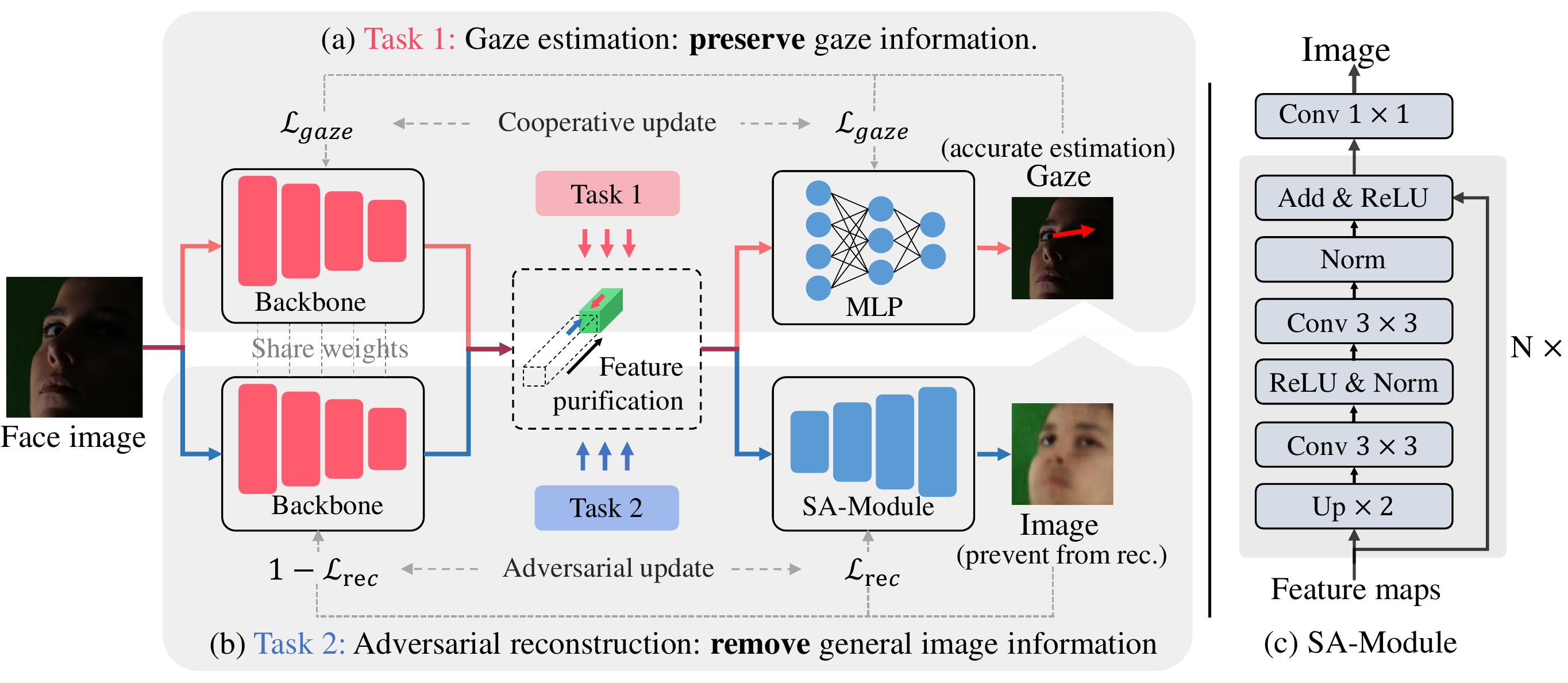}	
	\end{center}
	\caption{The architecture of PureGaze. It is consisted of two share-weight backbones (ResNet-18) for feature extraction, one two-layer MLP (Muti-layer Perception) for gaze estimation, and one SA-Module (N=5) for recovering images. The backbone and MLP are cooperative to preserve gaze information, \ie, perform gaze estimation, while the backbone and SA-Module are adversarial to remove general image information, \ie, perform adversarial reconstruction. The backbone simultaneously performs the two tasks while the two tasks are not cooperative but adversarial. 
	The backbone performs adversarial learning with itself  to purify extracted feature. }
	\label{fig:network}
\end{figure*}

As shown in \Fref{fig:pipeline}, the key idea of our method is to purify gaze feature,~\ie,~we eliminate gaze-irrelevant factors such as illumination and identity.
The purified feature is more generalized than original feature, and naturally brings improvement in cross-domain performance. 
To be specific, we propose a plug-and-play self-adversarial framework.
As shown in \Fref{fig:overview}, the framework contains two tasks, which are to preserve gaze information and to remove general facial image information.
Simultaneously optimizing the two tasks, we implicitly purify the gaze feature without defining gaze-irrelevant feature.
In fact, it is also non-trivial to define all gaze-irrelevant features. 
We also realize the framework with a practical neural network.
As shown in ~\Fref{fig:network}, the two tasks are respectively approximated as a gaze estimation task and an adversarial reconstruction task.
We propose the final PureGaze to simultaneously perform the two tasks to purified the gaze feature.
The PureGaze contains a plug-and-play SA-Module, which can be used to enhance existing gaze estimation methods directly and significantly.

The contributions of this work are threefold:
\begin{itemize}
	\item We propose a plug-and-play domain-generalization framework for gaze estimation methods. It improves the cross-dataset performance without knowning the target dataset or touching any new samples. To the best of our knowledge, it is the first domain-generalization framework in gaze estimation.
	
	\item The domain-generalizability comes from our proposed gaze feature purification. We design a self-adversarial framework to purify  gaze features, which eliminates the gaze-irrelevant factors such as illumination and identity. The purification is easily explainable via visualization as shown in the experiment.
	
	\item Our method achieves state-of-the-art performance in many benchmarks. Our plug-and-play module also enhances existing gaze estimation methods significantly.
	
\end{itemize}

\section{Related Works}
\textbf{Typical Gaze Estimation.} 
Recently, many gaze estimation methods are proposed.
Cheng~\etal explore the two-eye asymmetry ~\cite{Cheng_2018_ECCV,Cheng_2020_tip}.
Park~\etal generate pictorial gaze representation to handle subject variance~\cite{Park_2018_ECCV}.
Fisher~\etal leverage two VGG networks to process two eye images~\cite{Fischer_2018_ECCV}.
Zhang~\etal utilize attention mechanism to weight facial feature ~\cite{Zhang_2017_CVPRW}.
Chen~\etal leverage dilated convolution to estimate gaze~\cite{Chen_2019_ACCV}.
Bao~\etal leverage face and eye images to estimate point of gaze~\cite{Bao_2020_ICPR}.
Zheng~\etal propose a gaze/head redirection network and use generated images for data augmentation~\cite{zheng2020self}.
Cheng~\etal estimate gaze from facial images and refine the gaze with eye images~\cite{Cheng_2020_AAAI}.

\noindent\textbf{Cross-domain Gaze Estimation.}
Zhang~\etal~\cite{Zhang_2018_CHI} fine tune the pre-trained model in target domain. 
Wang~\etal~\cite{Wang2_2019_CVPR} and Kellnhoder~\etal ~\cite{Kellnhofer_2019_ICCV} propose to use adversarial learning to align the features in the source and target domain.
Liu~\etal~\cite{liu2021generalizing} propose an ensemble of networks that learn collaboratively with the guidance of outliers. 
These methods utilize data from target domain, which is not always user-friendly.

\section{Overview}
\subsection{Definition of the Purification}
We first formulate the proposed self-adversarial framework in this section.
Without loss of generality, we formulate the gaze estimation problem as
\begin{equation}
	\label{equ:basicgaze}
	\bv{g} = F_\phi(E_\theta(\bm{I})),
\end{equation}
where $E_\theta$ is a feature extraction function (\eg~neural networks), $F_\phi$ is a regression function, $\bm{I}$ is a face/eye appearance and $\bv{g}$ is a estimated gaze.
We use $\bm{Z}$ to denote the extracted feature,~\ie,~$\bm{Z} = E_\theta(\bm{I})$.

We slightly abuse the notation $I$ to represent a set of all features in one image.
We can simply divide the whole image feature into two subsets, gaze-relevant feature $G$ and gaze-irrelevant feature $N$.
It is easy to get the relation:

\begin{equation}
	\label{equ:relation}
	G \cup N= I \quad and \quad  G\cap N =\varnothing
\end{equation}

Our goal is to find the optimal $E_{\theta^*}$ to extract purified feature
$Z^*$, where $Z^*$ does not contain gaze-irrelevant feature,~\ie,  $Z^* \cap N =\varnothing$. 
Besides, we believe the feature which has weak relations with gaze should be also eliminated to improve generalization. 

\subsection{Self-Adversarial Framework}
\label{sec:framework}
As shown in \Fref{fig:overview}, we design two tasks for feature purification.
The first task is to minimize the mutual information (MI) between image feature and the extracted feature,~\ie,
\begin{equation}
	\label{equ:min}
	\theta^* = \mathop{\arg\min}_{\theta} H(I, Z)
\end{equation}
The function $H(X, Y)$ computes the MI between $X$ and $Y$.
It indicates the relation between $X$ and $Y$,~\eg~, $H(X, Y) = 0$ if $X$ is independent with $Y$.
This task also means the extracted feature should contain less image information. 

The other task is to maximize the MI between gaze-relevant feature and extracted feature,~\ie,
\begin{equation}
	\label{equ:max}
	\theta^* = \mathop{\arg\max}_{\theta} H(G, Z),
\end{equation}
This constraint means the extracted feature should contain more gaze-relevant information.

\subsection{Learning to Purify in the Framework}
We simultaneously solve \Eref{equ:min} and \Eref{equ:max}.
In other words, the extracted feature needs to contain more gaze information (\Eref{equ:max}) and less image information (\Eref{equ:min}).
The two optimization tasks compose a self-adversarial framework on the extracted feature.
During the optimization, gaze-irrelevant feature is eliminated to satisfy~\Eref{equ:min} and gaze-relevant feature is preserved to satisfy~\Eref{equ:max}.
In the other word, we purify extracted feature with the self-adversarial framework.

In addition, \Eref{equ:min} and \Eref{equ:max} implicate the minimax problem of $H(G, Z)$.
It is intuitive that the extracted feature will gradually discard some gaze-relevant information to decrease image information,~\ie, to satisfy \Eref{equ:min}.
Meanwhile, to satisfy \Eref{equ:max}, the feature having weak relations with gaze will be discarded first.

\section{PureGaze}

In the previous section, we propose two key tasks, \ie, \Eref{equ:min} and ~\Eref{equ:max}. The two tasks compose a self-adversarial framework to purify feature. In this section, we propose PureGaze based on the framework. 
We realize the two tasks with two practical tasks, gaze estimation and adversarial reconstruction.\footnote{The detailed deduction of realization is shown in the Supplementary Material.} We also propose two loss function for the framework.

\textbf{Gaze estimation:} 
We use gaze estimation tasks to preserve gaze information in the extracted feature,~\ie, \Eref{equ:max}.
In fact, the task can be realized with any gaze estimation network.
We simply divide gaze estimation networks into two subnets, backbone for extracting feature and MLP for regressing gaze from the feature (\Fref{fig:network}(a)).
We use a gaze loss function $\mathcal{L}_{gaze}$ such as L1 loss to optimize the two subnets.
The two subnets cooperate to preserve gaze information.

\noindent\textbf{Adversarial reconstruction:}
We propose an adversarial reconstruction task to remove general image information from extracted feature,~\ie,~\Eref{equ:min}.
Our assumption is that if the reconstruction network cannot recover input images from extracted feature, it means the extracted feature contains no image information.

Therefore, we first propose a SA-Module for reconstruction as shown in ~\Fref{fig:network}(c).
It contains a block for upsampling and a $1\times1$ convolution layer to align channels.
Further, the network architecture for adversarial reconstruction is shown in ~\Fref{fig:network}(b).
We use a backbone for feature extraction and SA-Module for recovering images.
We assign adversarial losses to the backbone and SA-Module.
The SA-Module tires to reconstruct images and is optimized with an reconstruction loss $\mathcal{L}_{rec}$ such as pixel-wise MSE Loss.
The backbone tires to prevent the reconstruction.
We use an adversarial loss $\mathcal{L}_{adv}$ to optimize it, where
\begin{equation}
	\mathcal{L}_{adv}=1-\mathcal{L}_{rec}.
\end{equation}

It is obvious that the backbone and the SA-Module are adversarial in reconstruction, ~\ie, the backbone finally removes general image information from  extracted feature.

\subsection{Architecture of PureGaze}
The architecture of PureGaze is shown in the left part of ~\Fref{fig:network}.
We respectively build two networks for gaze estimation and adversarial reconstruction with the same backbone, and share the weight of two backbones.  

In general, PureGaze contains three networks, which are a backbone for feature extraction, a MLP for gaze estimation and a SA-Module for image reconstruction.
The loss functions of the three parts are
\begin{equation}
	\mathcal{L}_{SA}= \mathcal{L}_{rec}.
\end{equation}
\begin{equation}
	\mathcal{L}_{MLP}= \mathcal{L}_{gaze}.
\end{equation}
\begin{equation}
	\mathcal{L}_{backbone}= \alpha\mathcal{L}_{adv} + \beta\mathcal{L}_{gaze}.
\end{equation}
where $\alpha$ and $\beta$ are hyper-parameters. 
In this paper, we use L1 Loss for gaze estimation and pixel-wise MSE for reconstruction:
\begin{equation}
	\mathcal{L}_{gaze}= \left\| \bv{g} -  \hat{\bv{g}} \right\|_1.
\end{equation}
\begin{equation}
	\mathcal{L}_{rec}= \left\| I -  \hat{I} \right\|_2.
\end{equation}

\noindent\textbf{Purifying Feature in Training:}
PureGaze uses one backbone to extract feature.
The backbone has two goals, minimizing $\mathcal{L}_{gaze}$ and minimizing $\mathcal{L}_{adv}$. 
Minimizing $\mathcal{L}_{gaze}$ means the backbone should extract gaze-irrelevant feature, while minimizing $\mathcal{L}_{adv}$ means the backbone should not extract any image feature.
The two goals are not cooperative but adversarial, and compose an adversarial learning to purify the extract feature.
In addition, $\mathcal{L}_{adv}$ is easily satisfied with learning a local optimal solution to cheat the SA-Module.
We design another task $\mathcal{L}_{rec}$ to against $\mathcal{L}_{adv}$ to avoid the local optimal solution.
The two novel adversarial tasks both are important parts in PureGaze. 
\begin{table*}[t]
	\renewcommand\arraystretch{1.3}
	\setlength\tabcolsep{12pt}
	\normalsize
	\caption{Performance comparison with SOTA methods. PureGaze shows best performance among typical gaze estimation methods (\textit{w/o} adaption), and has competitive result among domain adaption methods. Note that, PureGaze learns one optimal model for four tasks, while domain adaption methods need to learn a total of four models. This is an advantage of our method.}
	\begin{tabular}{p{1.3cm}|p{5.3cm}|c|c|c|c|c}
		
		\toprule
		\multirow{2}{*}{Category}&\multirow{2}{*}{Methods} &Target & \multirow{2}{*}{G$\rightarrow$M}& \multirow{2}{*}{G$\rightarrow$D}&\multirow{2}{*}{E$\rightarrow$M}& \multirow{2}{*}{E$\rightarrow$D} \\
		
		&&Samples&&&\\
		\hline
		\multirow{7}{*}{\tabincell{c}{without\\ adaption}}& RT-Gene~\cite{Fischer_2018_ECCV}&\NA&$21.81^\circ$ &$38.60^\circ$&-&-\\
		&Dilated-Net~\cite{Chen_2019_ACCV}&\NA& $18.45^\circ$&  $23.88^\circ$&-&-\\
		&Full-Face~\cite{Zhang_2017_CVPRW}&\NA&$11.13^\circ$& $14.42^\circ$ &$12.35^\circ$ &$30.15^\circ$\\
		&CA-Net~\cite{Cheng_2020_AAAI}&\NA&$27.13^\circ$ &$31.41^\circ$ &-&-\\
		&ADL$^*$~\cite{Kellnhofer_2019_ICCV}&\NA&$11.36^\circ$& $11.86^\circ$& $7.23^\circ$&$8.02^\circ$ \\
		\cline{2-7}
		&Baseline (ours) &\NA&$9.89^\circ$&$11.42^\circ$&$8.13^\circ$&$7.74^\circ$\\
		&\textbf{PureGaze (ours)} &\NA& $\bm{9.28^\circ}$& $\bm{9.32^\circ}$ & $\bm{7.08^\circ}$&$\bm{7.48^\circ}$ \\
		
		\hline
		\hline
		\multirow{8}{*}{\tabincell{c}{with \\ adaption}}
		&ADL~\cite{Kellnhofer_2019_ICCV}&$>100$&{$9.70^\circ$}&{$10.28^\circ$}&{$5.48^\circ$}&{$16.11^\circ$}\\
		&DAGEN ~\cite{guo_2020_ACCV} &$\sim500$ & $6.61^\circ$& $12.90^\circ$&$6.16^\circ$&$9.73^\circ$\\
		&ADDA~\cite{tzeng2017adversarial}&$\sim500$&$8.76^\circ$&$14.80^\circ$&$6.33^\circ$&$7.90^\circ$\\
		&GVBGD~\cite{cui2020gradually}&$\sim1000$&$7.64^\circ$&$12.44^\circ$&$6.68^\circ$&$7.27^\circ$\\
		&UMA~\cite{cai2020generalizing}&$\sim100$&$8.51^\circ$&$19.32^\circ$&$7.52^\circ$&$12.37^\circ$\\
		&PNP-GA~\cite{liu2021generalizing}&$<100$&$6.18^\circ$&$7.92^\circ$&$5.53^\circ$&$\bm{5.87^\circ}$\\
		\cline{2-7}
		&Fine-tuned Baseline &$<100$&$5.28^\circ$&$7.66^\circ$&$5.68^\circ$&$7.26^\circ$\\
		&Fine-tuned PureGaze &$<100$& $\bm{5.20^\circ}$ & $\bm{7.36^\circ}$&$\bm{5.30^\circ}$&$6.42^\circ$
		
		\\
		\bottomrule
	\end{tabular}
	\label{table:Crossdataset}
\end{table*}

\subsection{Local Purification Loss} 
It is intuitive that eye region is more important than other facial regions for gaze estimation.
Therefore, we want PureGaze to pay more attention to purify the feature of eye region.
One simple solution is to directly use eye images as input.
However, we believe it is unreasonable since other facial regions also provide  useful information. 

As a result, we propose the local purification loss (LP-Loss). We use an attention map to focus the purification on a local region.
Note that, the attention map is only applied to $\mathcal{L}_{adv}$,~\ie, the loss function of the backbone is modified as 
\begin{equation}
	\mathcal{L}_{backbone} = \alpha \mathbb{E}[\bm{M}*(\bv{1}- (I_i -  \hat{I}_i)^2)] + \beta \mathcal{L}_{gaze}.
\end{equation}
where $\bm{M}$ is the attention map, and $*$ means element-wise multiplication. 
In this paper, we use mixed guassian distribution to generate the attention map.
We use the coordinates of two eye centers as mean values, and
the variance $\Sigma=diag(\sigma^2, \sigma^2)$ of the distribution can be customized.

The loss function can be understood as following.
On the one hand, LP-Loss focuses the purification on eye region. 
On the other hand, LP-Loss does not change the loss function of gaze estimation, \ie, $\mathcal{L}_{gaze}$.
Human gaze is estimated from whole face images rather than weighted face images.

\subsection{Truncated Adversarial Loss}
The adversarial reconstruction task plays an important role in the PureGaze.
It ensures the extracted feature contains less image feature.
In PureGaze, we minimize $\mathcal{L}_{adv}$ to prevent the reconstruction.
A smaller value of $\mathcal{L}_{adv}$ indicates a larger pixel difference between the generated and original images.
However, we think it is redundant to produce a very large pixel difference.
The reason is that $\mathcal{L}_{adv}$ is designed to prevent the reconstruction rather than recover an ``inverse" version of the original image.

Therefore, we further propose the truncated adversarial loss (TA-Loss).
We use a threshold $k$ to truncate the adversarial loss $\mathcal{L}_{adv}$. 
In one word, $\mathcal{L}_{adv}$ will be zero if the pixel difference is larger than $k$.
The final loss function of the backbone is:
\begin{equation}
	\mathcal{L}_{backbone} = \alpha \mathbb{E}[\bm{M}*\mathbbm{1}_{[\bv{1}- \left\| I_i -  \hat{I}_i \right\|_2 > k]}*(\bv{1}- (I_i -  \hat{I}_i)^2)] + \beta\mathcal{L}_{gaze}.  
\end{equation}
where $\mathbbm{1}$ is the indicator function and $k$ is the threshold.

\section{Experiments}
\subsection{Data-preprocessing }
\textbf{Task definitions:} We use Gaze360~\cite{Kellnhofer_2019_ICCV} and ETH-XGaze~\cite{Zhang_2020_ECCV} as training set, since they have a large number of subjects, various gaze range and head pose. 
We test our model in two popular datasets, which are MPIIGaze~\cite{Zhang_2017_CVPRW} and EyeDiap~\cite{Mora_2014_ETRA}.
We totally conduct four cross-dataset tasks, and denote them as E (ETH-XGaze)$\rightarrow$M (MPIIGaze),  E$\rightarrow$D (EyeDiap), G (Gaze360)$\rightarrow$M, G$\rightarrow$D.

\noindent\textbf{Data Preparing.}
We follow~\cite{Cheng_2021_arxiv} to prepare datasets.
Gaze360~\cite{Kellnhofer_2019_ICCV} dataset contains a total of 172K images from 238 subjects.
Note that some of the images in Gaze360 only captured the back side of the subject. 
These images is not suitable for appearance-based methods.
Therefore, we first clean the dataset with a simple rule.
We remove the images without face detection results based on the provided face detection annotation. 
ETH-XGaze~\cite{Zhang_2020_ECCV} contains a total of 1.1M images from 110 subjects.
It provides a training set containing 80 subjects.
We split 5 subjects for validation and others are used for training.
MPIIGaze~\cite{Zhang_2017_CVPRW} is prepared based on the standard protocol.
We collect a total of 45K images from 15 subjects.
EyeDiap~\cite{Mora_2014_ETRA} provide a total of 94 video clips from 16 subjects.
We follow the common steps to prepare the data as in ~\cite{Zhang_2017_CVPRW,Cheng_2020_AAAI}.
Concretely, we select the VGA videos of screen targets session and sample one image every fifteen frames.
We also truncate  the data to ensure the number of images from each subject is the same.

\noindent\textbf{Data rectification.}
Data rectification is performed to simplify the gaze estimation task.
We follow \cite{Sugano_2014_CVPR} to process MPIIGaze and \cite{Zhang_2018_etra} to process EyeDiap.
ETH-XGaze is already rectified before publication.
Gaze360 rectifies their gaze directions to cancel the effect caused by camera pose.
We directly use their provided data.

\subsection{Comparison Methods}
\textbf{Baseline}: We remove the SA-Module in PureGaze. The new network is denoted as Baseline. It is obvious the performance difference between PureGaze and Baseline is caused by SA-Module.
We also denote the feature extracted by Baseline as original feature, and the feature extracted by PureGaze as purified feature.

\noindent\textbf{Gaze estimation methods}: 
We compare our method with four methods, which are Full-Face~\cite{Zhang_2017_CVPRW}, RT-Gene~\cite{Fischer_2018_ECCV}, Dilated-Net~\cite{Chen_2019_ACCV} and CA-Net~\cite{Cheng_2020_AAAI}.
These methods all perform well in within-dataset evaluation.
We implement Full-Face and Dilated-Net using Pytorch,
and use the official code of the other two methods.

\noindent\textbf{Domain adaption methods}: 
We also compare our method with domain adaption methods for reference.
In fact, it is unfair to compare our method with domain adaption methods since these methods require target samples.
Adversarial learning (ADL)~\cite{Kellnhofer_2019_ICCV} is proved useful in gaze estimation, and has a similar feature with our method. 
We implement ADL for main comparison. 
We also modify the method as ADL$^*$, where we only use a discriminator to distinguish personal feature in source domains.
ADL$^*$ does not require target samples as our methods.
In addition, we directly report the performance of other domain adaption methods from~\cite{liu2021generalizing} for reference.

\begin{table}[t]
	\renewcommand\arraystretch{1.2}
	\setlength\tabcolsep{6pt}
	\normalsize
	\caption{We apply the self-adversarial framework into other advanced gaze estimation methods. Our framework directly enhances existing gaze estimation methods. The experiment also provides a more fair comparison with these methods.}
	\begin{tabular}{l|c|c|c|c}
		\hline
		Methods &G$\rightarrow$M& G$\rightarrow$D &E$\rightarrow$M & E$\rightarrow$D \\
		\hline
		\hline
		Full-Face &$11.13^\circ$&$14.42$ &$12.35^\circ$ &$30.15^\circ$ \\
		Full-Face+SA (ours) &$9.16^\circ$&$14.20$  &$11.50^\circ$ &$21.01^\circ$  \\
		\hline
		CA-Net &$27.13^\circ$ &$31.41^\circ$ &-&- \\
		CA-Net+SA (ours)&$\bm{9.03^\circ}$ &$9.71^\circ$ &-&-\\
		\hline
		Baseline (ours)&$9.89^\circ$&$11.42^\circ$ & $8.13^\circ$&$7.74^\circ$\\
		PureGaze (ours) & $9.28^\circ$& $\bm{9.32^\circ}$ & $\bm{7.08^\circ}$&$\bm{7.48^\circ}$  \\
		\hline
	\end{tabular}
	\label{table:pnp}
\end{table}

\subsection{Performance Comparison with SOTA Methods}
\label{A}
We first conduct experiments in four cross-dataset tasks.
The result is shown in \Tref{table:Crossdataset}.
Note that, Dilated-Net, CA-Net and RT-Gene are not applicable in ETH-XGaze, since ETH-XGaze cannot always provide reliable eye images.
In addition, ETH-XGaze dataset uses an off-the-shelf ResNet50 as baseline.
We follow the protocol and replace the backbone in our method and ADL with ResNet50 in ETH-XGaze.

\noindent\textbf{Comparison with typical gaze estimation methods:} The second row of \Tref{table:Crossdataset} shows the comparison between our method and gaze estimation methods.
Our method and compared methods are all trained on source domains and evaluated on target domains.  
It is obvious current gaze estimation methods usually have bad performance in cross-dataset evaluation.
This is because these methods are easily over-fitted in source domains.
In contrast, our Baseline has good performance in all tasks due to simple architecture.
PureGaze further improves the performance of Baseline and achieves the state-of-the-art performance in all tasks.

\noindent\textbf{Comparison with domain adaption methods:}
The third row of Table 1 shows the performance of domain adaption methods. 
ADL has the same backbone as PureGaze.
It improves the performance in three tasks and fails in $E\rightarrow D$.
It also has the best performance among compared methods in $E\rightarrow M$.
Compared with ADL, PureGaze surpasses the performance of ADL in three tasks without target samples.
This proves the effectiveness of our method.

On the other hand, we evaluate the performance of ADL$^*$ and show the result in the second row of Table 1.
Without target samples, ADL$^*$ cannot always bring performance improvement.
This is because it is hard to improve performance in all unknown domains, and also   demonstrate the value of our method. PureGaze surpasses ADL $^*$ in all tasks.

Table 1 also shows the performance of other domain adaption methods.
PureGaze shows competitive result among these domain adaption methods without domain adaption.
In addition, we also provide a simple application of PureGaze, where we sample 5 images per person from target domains to fine tune PureGaze.
Fine-tuned PureGaze further improves the performance with a fast calibration/adaption.
More valuable, we observe fine-tuned PureGaze also has better performance than fine-tuned Baseline.
This proves PureGaze learns a better feature representation.

\begin{figure}[t]
	\begin{center}
		\subfigure[Training on Gaze360]{
			\label{fig:diffK_gaze360}
			\includegraphics[width=0.45\columnwidth]{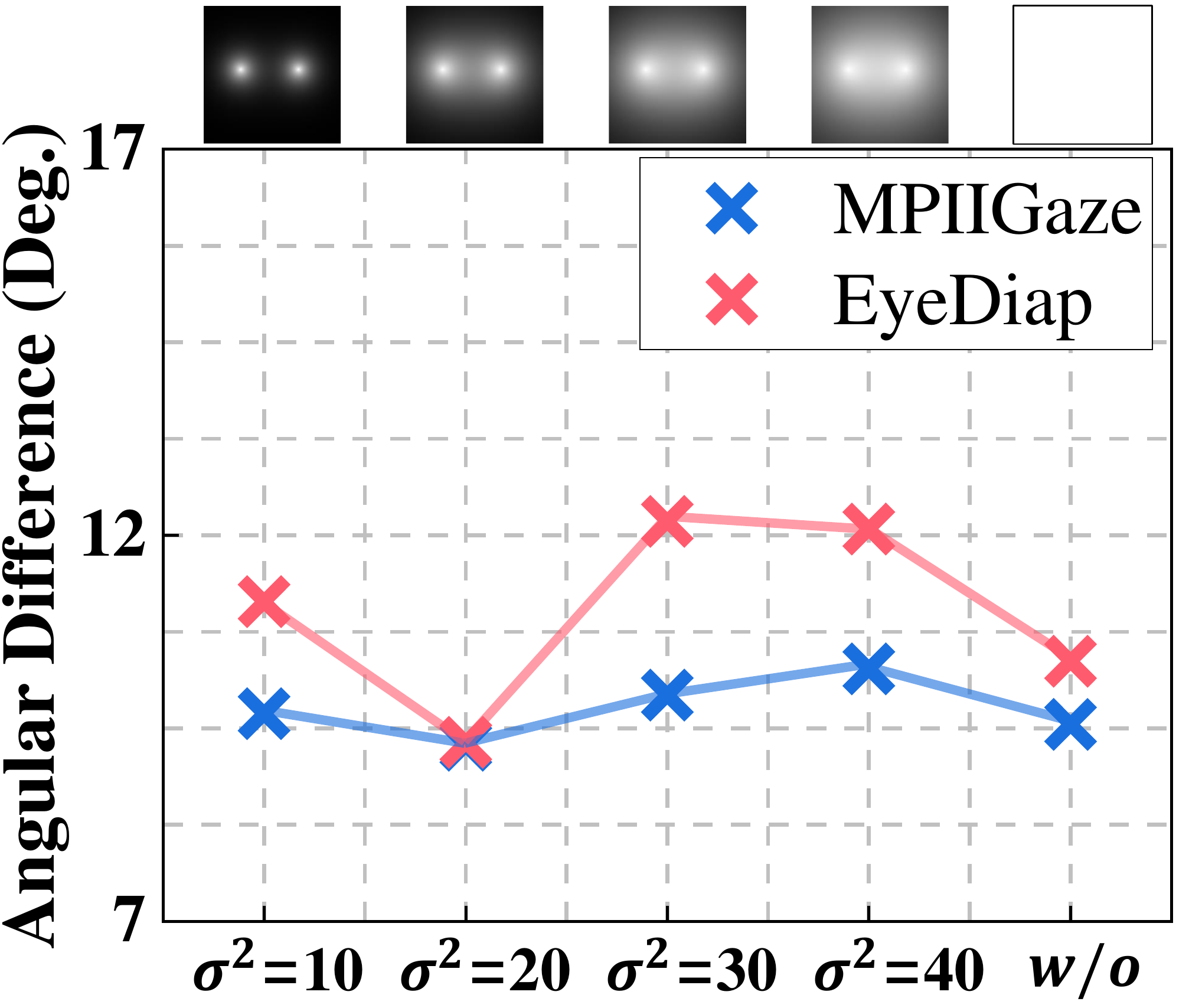}
		}
		\subfigure[Training on ETH]{
			\label{fig:diffK_eth}
			\includegraphics[width=0.45\columnwidth]{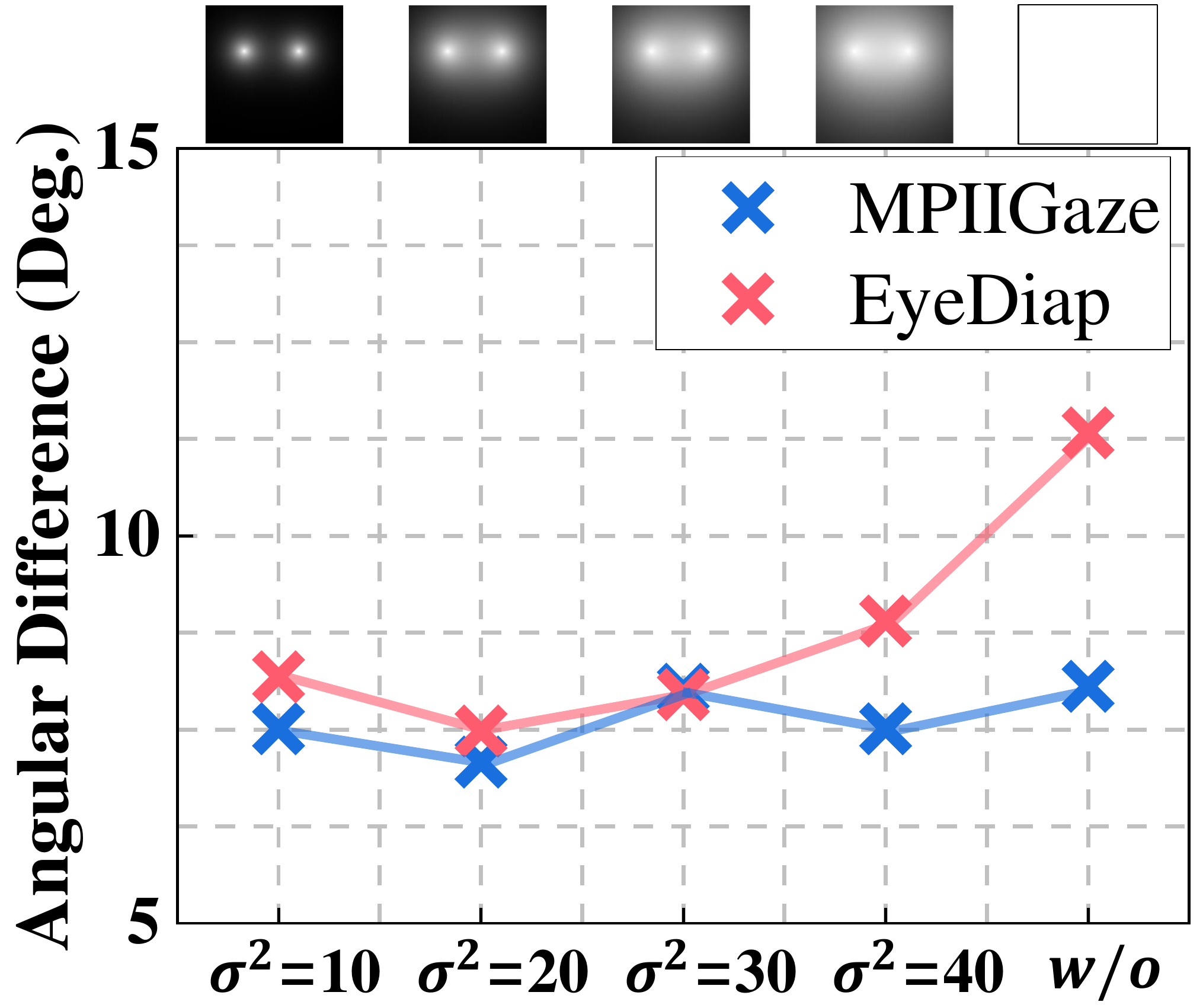}
		}
		
		\subfigure[Training on Gaze360]{
			\label{fig:diffT_gaze360}
			\includegraphics[width=0.45\columnwidth]{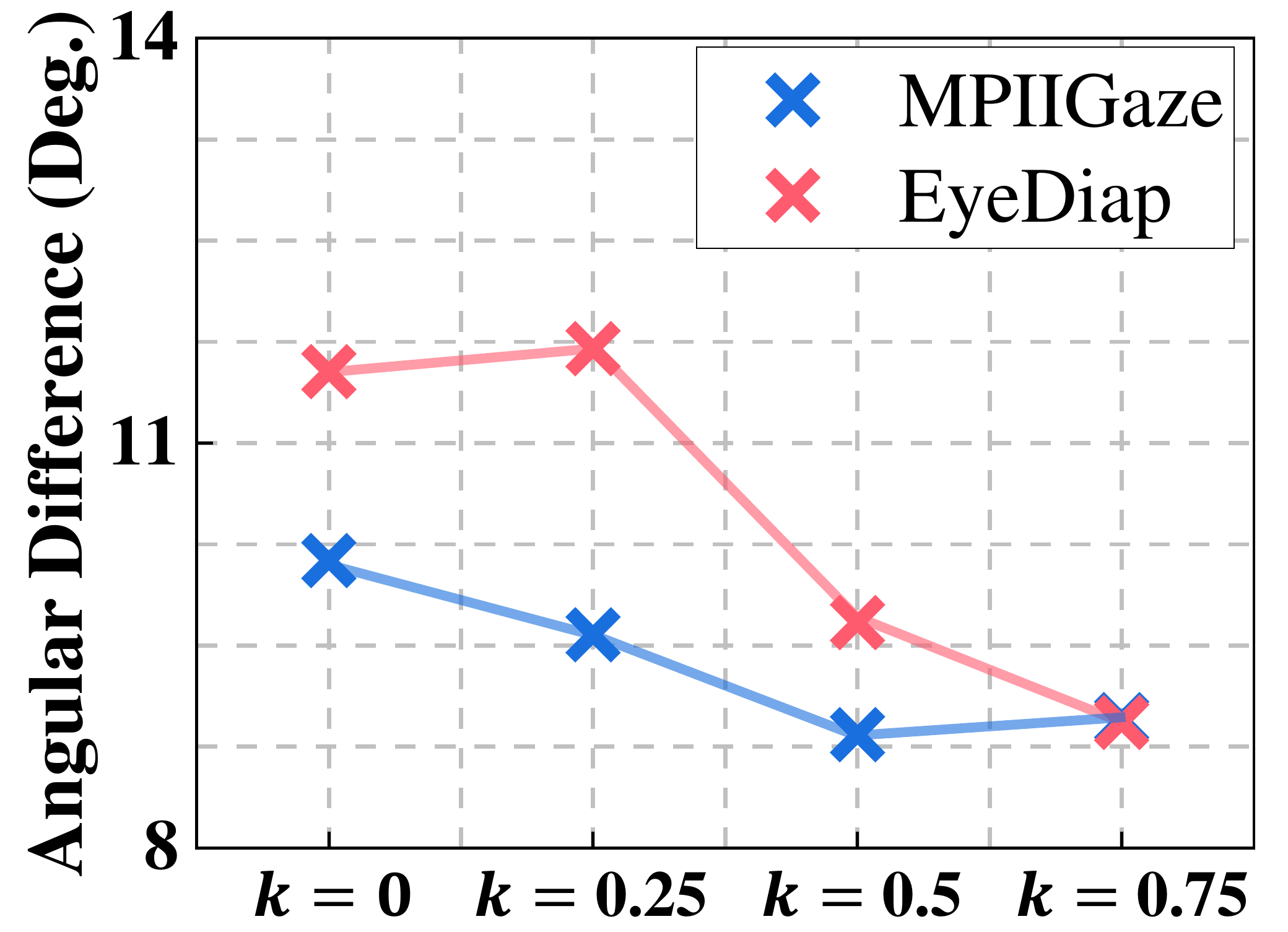}
		}
		\subfigure[Training on ETH]{
			\label{fig:diffT_ETH}
			\includegraphics[width=0.45\columnwidth]{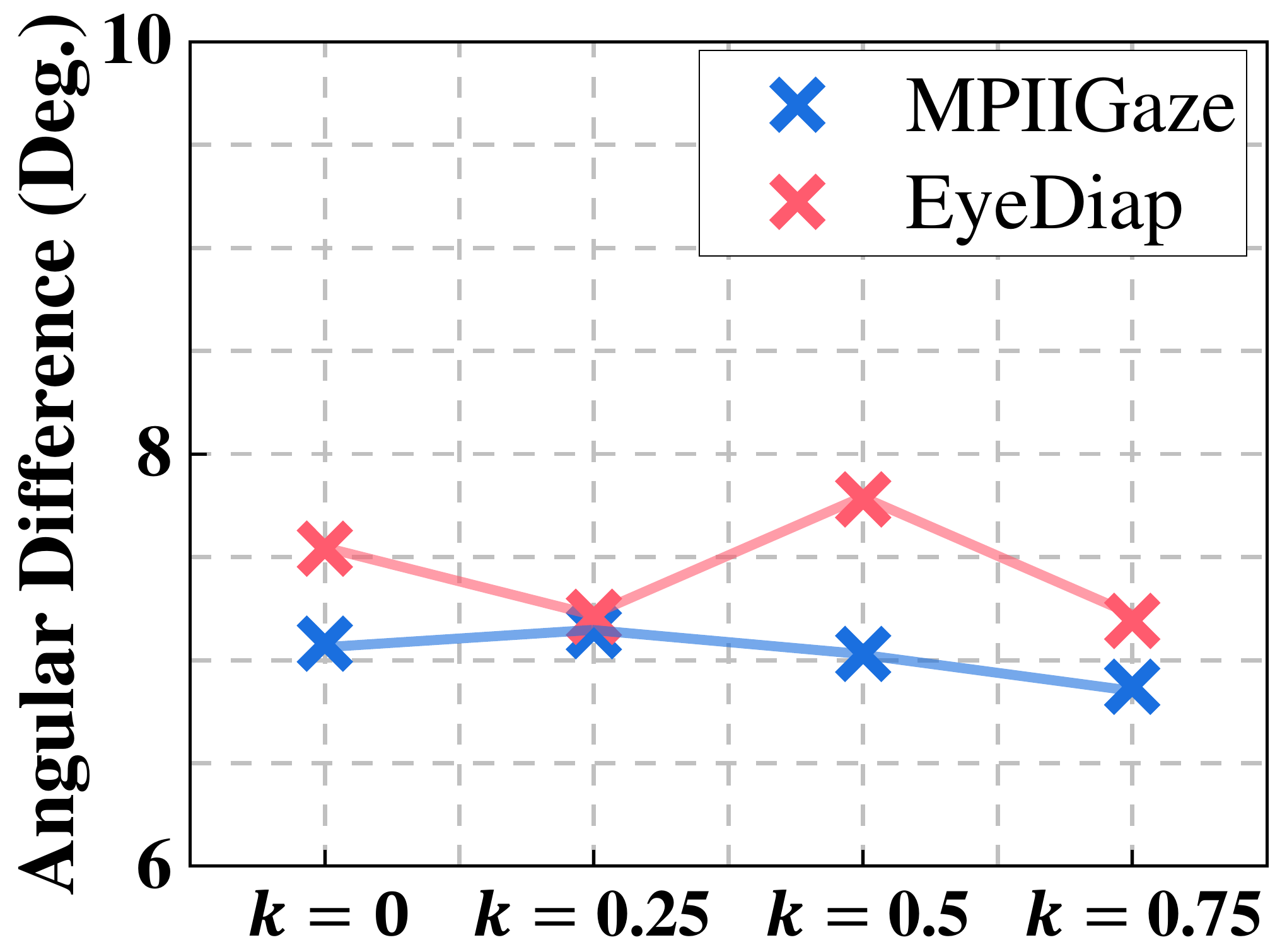}
		}
	\end{center}
	
	\caption{We evaluate the hyper-parameters of the two loss function. The first row shows the result about LP-Loss, where $w/o$ means we ablate the loss. The second row shows the result about TA-Loss, where $k=0$ means we ablate TA-Loss. Our method is the best when $\sigma^2$ is 20 and $k$ is 0.75.}
	
	\label{fig:ablationKS}
	
\end{figure}
\begin{figure*}[t]
	\begin{center}
		\subfigure[Reconstruction by purified feature: less identity \& more accurate gaze information.]{
			\label{fig:visualA}
			\includegraphics[width=1\columnwidth]{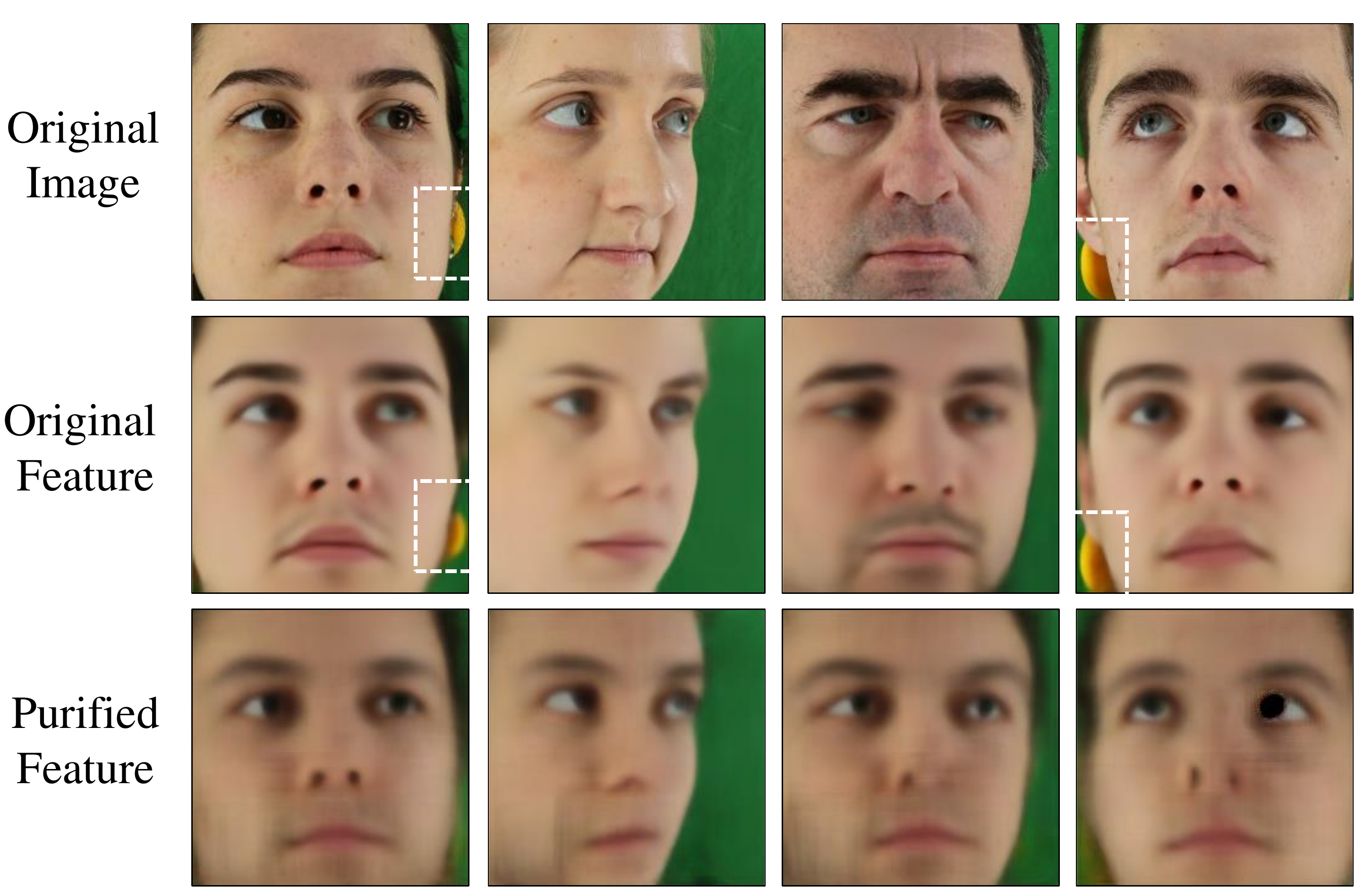}
		}
		\subfigure[Reconstruction by purified feature: less affected by illumination.]{
			\label{fig:visualB}
			\includegraphics[width=1\columnwidth]{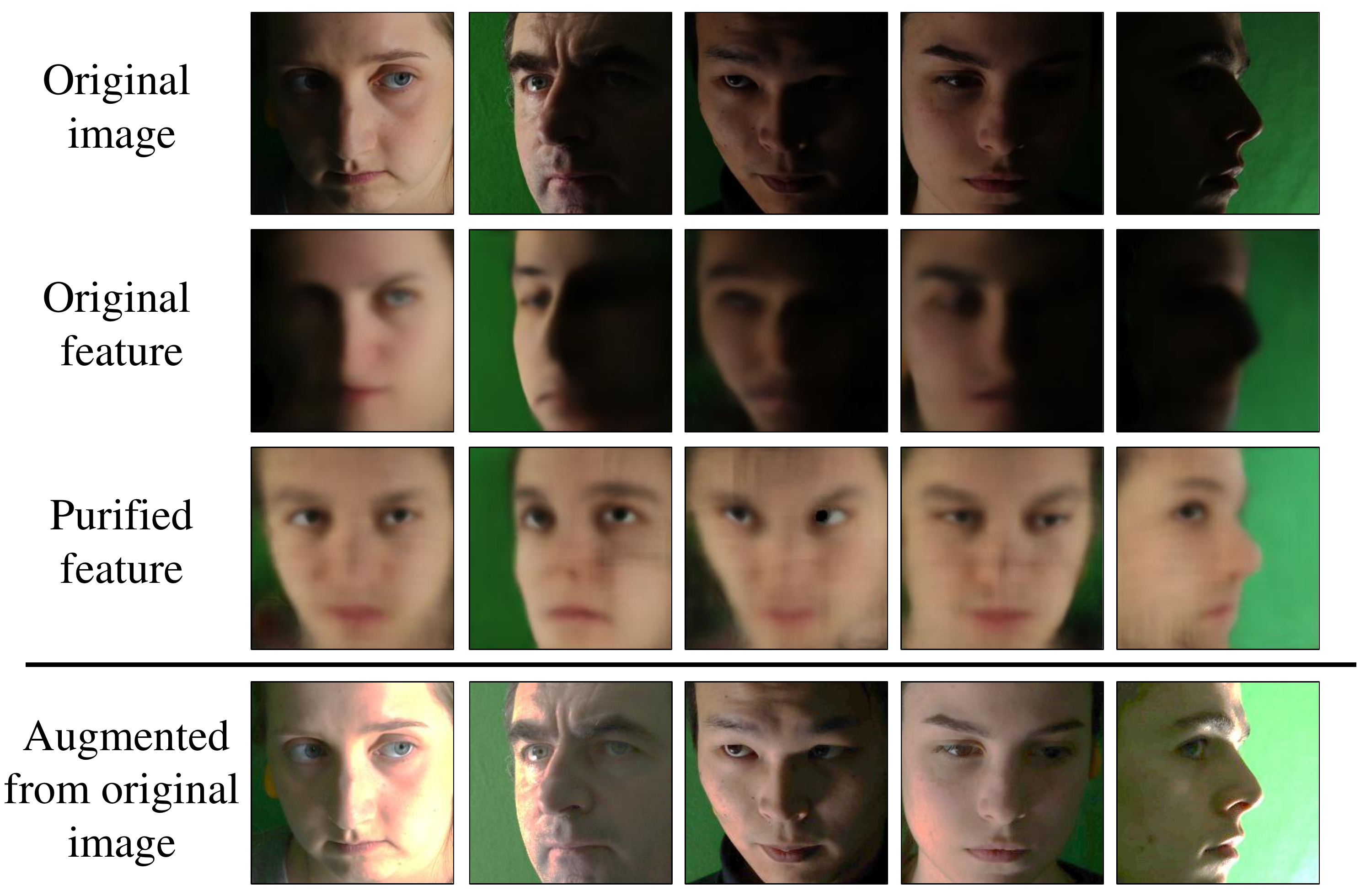}
		}
	\end{center}
	\vspace{-1mm}
	\caption{We visualize the purified feature and the original feature via reconstruction. The result clearly explains the purification. Our purified feature contains less gaze-irrelevant feature and naturally improves the cross-domain performance.
	a) The purified feature contains less identity information than the original feature.
	Besides, it is interesting that a head rest is captured in the first and fourth columns. The original feature also contains the head rest information while our method eliminates it. b) The purified feature contains less illumination information. We also manually augment each original image in the fourth row. The result shows our method accurately captures the gaze information under the dash area.}
	\label{fig:visualization}
\end{figure*}

\subsection{Plug Existing Gaze Estimation Methods}
We also apply our self-adversarial framework into Full-Face~\cite{Zhang_2017_CVPRW} and CA-Net~\cite{Cheng_2020_AAAI}.
We input their final facial feature maps into SA-Module, and simply add two loss functions, $\mathcal{L}_{rec}$ and $\mathcal{L}_{adv}$.

The result is shown in ~\Tref{table:pnp}.
It is surprising that CA-Net has the worst performance in  G$\rightarrow$M , while CA-Net+SA has the best performance in  G$\rightarrow$M.
Besides, it also improved by nearly $70\%$  in G$\rightarrow$D.
Full-Face+SA also shows better performance than Full-Face in all tasks. 
The experiment provides a more fair comparison with typical gaze estimation methods, and proves the plug-and-play attribute of our self-adversarial framework.
Note that, our framework does not require additional inference parameters and training images. 
This is a key advantage of our method.

\subsection{Ablation Study of Two Loss Function}
Self-adversarial framework provides a primary architecture of PureGaze while it is rough.
We also propose two loss functions (LP-Loss and TA-Loss) to enhance PureGaze.
The two loss functions both have hyper-parameters,~\ie~, variance $\sigma^2$ in LP-Loss and threshold $k$ in TA-Loss.
We conduct experiments about different hyper-parameters in this section.

As shown in \Fref{fig:diffK_gaze360} and \Fref{fig:diffK_eth}, we set four values for $\sigma^2$, which are $10$, $20$, $30$ and $40$, and also evaluate the performance without the loss.
We illustrate the generated attention maps in the top of the two figures.
As for TA-Loss, we set four values for $k$, which are $0$, $0.25$, $0.5$ and $0.75$. $k=0$ also means we ablate TA-Loss.
The results are shown in \Fref{fig:diffT_gaze360} and \Fref{fig:diffT_ETH}.
It is obvious that the two loss functions both brings performance improvement.
When $k=0.75$ and $\sigma^2=20$, PureGaze has best performance.

\subsection{Visualize Extracted Feature via Reconstruction}
To verify the key idea of gaze feature purification, we visualize purified features for further understanding. 
We provide reconstruction results of purified features and original features for comparison. We directly show the output of SA-Module to visualize the purified feature.
We freeze the parameters of the pre-trained model and simply train a SA-Module to reconstruct images from the original feature. 

According to the visualization result shown in \Fref{fig:visualization}, we could easily draw following conclusions:

\begin{itemize}
	\setlength{\itemsep}{5pt}
	\setlength{\parsep}{0pt}
	\setlength{\parskip}{0pt}
	\item[-] The purified feature contains less identity information than original feature.
	The reconstructed face appearances are approximately the same for each subject.
	
	\item[-] The purified feature contains less illumination factors.
	Besides, it is interesting that our method also recover a bright gaze region accurately from low-light images.
	This means our method is able to effectively extract gaze information under the dash area.
	
	\item[-] Except illumination and identity factors, our method also eliminates other gaze-irrelevant features like the head rest in ~\Fref{fig:visualA}.
	
\end{itemize}
Note that our method does not specify eliminated factors.
PureGaze automatically purifies the learned feature. 
This is an advantage of our method since it is non-trivial to manually list all gaze-irrelevant feature.

\section{Discussion}

\noindent\textit{1) Domain generalization.} 
Gaze estimation methods usually have large performance drop when tested in new environment.
This feature limits the application of gaze estimation.
In this paper, we innovate a new direction to solve the cross-domain problem. 
Compared with domain adaption (DA) methods, domain generalization (DG) methods are more flexible,~\eg, the setup of DA methods usually is time-consuming while DG methods can be directly applied to new domains.
But as a trade-off, DG methods usually perform worse than DA methods due to the lack of target domains information.
The trade-off between flexibility and accuracy should be considered by researchers. 

\noindent\textit{2) Self-adversarial framework.} 
We propose a self-adversarial framework to learn purified feature.
The purified feature improves the cross-domain performance without touching target samples.
In fact, our framework can also be considered as a zero-shot cross-domain method since we require no samples in target domains.
The zero-shot mechanism is designed based on the observation, gaze pattern is similar in all domains, while some gaze-irrelevant factors are usually domain-specific and bring performance drop.
Our method eliminates some gaze-irrelevant feature and naturally improves the cross-domain performance.
However, the same as DA methods, our method is unstable in source domains.
Our method slightly changes the performance in source domain ($\pm0.2^\circ$).
This is because an over-fitting model can achieve better performance in source domain compared with PureGaze.
Learning more generalized model is a future direction of our framework.

\section{Conclusion}
In this paper, we innovate a new direction to solve the cross-dataset problem in gaze estimation.
We propose a plug-and-play domain-generalization framework.
The framework purifies gaze feature to improve the performance in unknown target domains without touching the domain.
Experiments show our method achieves state-of-the-art performance among typical gaze estimation methods and also has competitive result compared with domain adaption methods.

\newpage
\appendix

\section{Appendix}

In the supplementary document, we first provide a background of gaze estimation.
Then, we provide the deduction from self-adversarial framework to PureGaze and describe the implementation details of PureGaze.
Finally, we provide additional experiments.
We give an apple-to-apple comparison between different types of gaze estimation methods. This comparison will provide a deep understanding about the advantage and disadvantage of domain generalization methods.
We also provide the experiments about illumination factors.
In the visualization, PureGaze eliminates illumination factors from extracted feature in very dark environment.
We further provide the quantitative improvement on each illumination intensity.
The quantitative result also matches the visualization result. 

\section{Background}
Appearance-based gaze estimation methods aim to infer human gaze $\bv{g}$ from appearance $\bm{I}\in\mathbb{R}^{H\times W \times C}$.
They usually learn a mapping function $\bv{g}=f(\bm{I})$ to directly map appearance to gaze.
Recently, Many methods leverage CNNs to model the mapping function and achieve outstanding performance~\cite{Wang_2019_CVPR,Wang2_2019_CVPR}.

Cross-person and cross-dataset problems are two key challenges in gaze estimation.
Conventional gaze estimation methods usually have large error in cross-person problem~\cite{Sugano_2014_CVPR}.
Recently, with the development of deep learning, appearance-based methods have good and saturated performance in cross-person problem.
To further improve the cross-person performance, researchers also aim to  learn person-specific models with few-shot additional calibration samples.
For example, Park~\etal propose a few-shot gaze estimation methods~\cite{Park_2019_ICCV}.
They first use an auto-encoder to learn feature representation and then use meta-learning to learn an person-specific gaze mapping from feature representation to gaze.
Yu~\etal propose a gaze redirection network~\cite{Yu_2019_CVPR}.
They use re-directed eye images to improve the performance in a specific user.

Cross-dataset problem is more challenging than cross-person problem.
Besides the personal difference, there are many known and unknown environment factors which have large impacts on cross-dataset performance.
Overview, we believe personal difference, environment difference and data distribution difference are three mainly problems in cross-dataset problem.
Our work provides an framework to handle the first two problems, which is a contribution of our work.

\textbf{Evaluation metric:}
We use angular error as the evaluation metric like most of methods~\cite{Cheng_2020_AAAI,Park_2019_ICCV,Zhang_2017_tpami}. 
Assuming the actual gaze direction is $\bm{g}\in\mathbb{R}^3$ and the estimated gaze direction is $\bm{\hat{g}}\in\mathbb{R}^3$, the angular error can be computed as:
\begin{equation}
\label{equ:3dmetric}
\mathcal{L}_{\mathrm{angular}} = \frac{\bm{g}\cdot\bm{\hat{g}}}{||\bm{g}||||\bm{\hat{g}}||}.
\end{equation}
A smaller error represents a better model. 

\section{Methodology}

\subsection{Deduction from the Framework to PureGaze}
In the main manuscript, we propose a self-adversarial framework containing two adversarial tasks:
\begin{equation}
\label{equ:min}
\theta^* = \mathop{\arg\min}_{\theta} H(I, Z)
\end{equation}
and
\begin{equation}
\label{equ:max}
\theta^* = \mathop{\arg\max}_{\theta} H(G, Z).
\end{equation}

Here, we introduce how to deduct the PureGaze from the two formulations.

To realize the framework, we first simplify \Eref{equ:min} and \Eref{equ:max}.
The mutual information $H(X, Y)$ can be further deduced as:
\begin{equation}
\label{equ:mut}
\begin{aligned}
H(X, Y) & = \mathbb{E}_{x\sim p(x),y\sim p(y)}[\log p(x|y)] - \mathbb{E}_{x\sim p(x)}[\log p(x)] \\
&\propto \mathbb{E}_{x\sim p(x),y\sim p(y)}[\log p(x|y)].
\end{aligned}
\end{equation}

Substituting \Eref{equ:mut} into \Eref{equ:min} and \Eref{equ:max}, we get:
\begin{equation}
\label{equ:min_new}
\theta^* = \mathop{\arg\min}_{\theta} \mathbb{E}_{i\sim p(I),z\sim p(Z)}[\log p(i|z)],
\end{equation}
and
\begin{equation}
\label{equ:max_new}
\theta^* = \mathop{\arg\max}_{\theta} \mathbb{E}_{g\sim p(G),z\sim p(Z)}[\log p(g|z)].
\end{equation}

\Eref{equ:min_new} means we should minimize the probability $p(i|z)$.
We approximate this probability with a reconstruction network $R: Z\rightarrow I$.
To minimize $p(i|z)$, we want $R$ to fail reconstructing images from the extracted feature.
In other words, the feature extraction network $E$ should extract the feature which is independent with images. 
Certainly, it is easy for feature extraction network $E$ to fool a pre-trained reconstruction network. And it is time-consuming to train a special reconstruction network for each iteration.
Thus, $E$ and $R$ are designed to perform an adversarial reconstruction task.
$R$ tries to reconstruct images from the feature,
while $E$ tries to prevent the reconstruction.
With the adversarial reconstruction, $E$ will discard all image information so that $R$ cannot reconstruct images,~\ie,~minimize $H(I, Z)$.

\Eref{equ:max_new} means we should maximize the probability of $p(g|z)$,~\ie,~ given image feature $z$, we should accurately recover gaze information from $z$.
We approximate $p(g|z)$ as a gaze regression network $F:Z\rightarrow G$.
$F$ aims to accurately estimate gaze from the extracted feature.

\begin{table*}[t]
	\renewcommand\arraystretch{1.4}
	\setlength\tabcolsep{5pt}
	\normalsize
	\caption{We provide an apple-to-apple comparison between different types of methods. We select three methods with the same backbone for fair comparison. PureGaze (domain generalization methods) achieves better performance than Baseline (typical gaze estimation methods) under the same condition. This is an advantage of our method. ADL (domain adaption methods) also has better performance than Baseline while it requires time for setup and target samples for adaption. }
	\begin{tabular}{c|c|cccc|c|c|c|c}
		\toprule
		\multirow{2}{*}{Methods}&\multirow{2}{*}{Types}   & \multicolumn{4}{c|}{Accuracy} & \multirow{2}{*}{Params} & \multirow{2}{*}{Setup} & \multirow{2}{*}{Target Samples} & \multirow{2}{*}{Model}\\
		\cline{3-6}
		&&G$\rightarrow$M& G$\rightarrow$D &E$\rightarrow$M & E$\rightarrow$D&&&&\\
		\midrule
		Baseline &Typical gaze estimation&$9.89^\circ$&$11.42^\circ$ & $8.13^\circ$&$7.74^\circ$&$11.2$M&Not required&Not required&Unique\\
		PureGaze &Domain generalization& $9.28^\circ$& $9.32^\circ$ & $7.08^\circ$&$7.48^\circ$&$11.2$M&Not required&Not required&Unique \\
		ADL&Domain adaption&{$9.70^\circ$}&{$10.28^\circ$}&{$5.48^\circ$}&{$16.11^\circ$}&$11.2$M&$>1$h& $> 50 $ images&Specific\\
		\bottomrule
	\end{tabular}
	\label{table:Crossdataset}
\end{table*}

\subsection{Implementation Detail}
In this paper, we use the convolutional part of ResNet-$18$ as the backbone.
We use a two-layer MLP for gaze estimation, where the output numbers of the two layers are $1000$ and $2$.
As for SA-Module, we use a five-layers SA-Module (N=5).  
The numbers of feature maps are 256, 128, 64, 32, 16 for each block (from bottom to top) and 3 for last $1\times1$ convolutional layer.
We use Adam for optimization and the learning rate is $0.0001$ for all three networks.
We empirically set $\alpha$ and $\beta$ as $1$ and $k$ as $0.75$. 
The $\sigma^2$ of attention map is $20$ pixel.
Indeed, the backbone, MLP and SA-Module are arbitrary where we only need to ensure the SA-Module can output a image having the same sizes with input image.

\section{Additional Experiments}

\subsection{Comparison between Different Types of Methods}
Our method is a domain generalization method.
We further provide an apple-to-apple comparison among our method, typical gaze estimation methods and domain adaption methods in this section.
We choose the Baseline as the representation of typical gaze estimation methods and ADL~\cite{Kellnhofer_2019_ICCV} as the representation of domain adaption methods.
The three methods have the same backbone and provide a fair comparison.

The result is shown in \Tref{table:Crossdataset}. 
PureGaze, Baseline and ADL have the same number in parameters.  
This is because PureGaze and ADL both are plug-and-play.
The two methods learn more generalized feature representation rather than changes the network architecture to improve cross-domain performance. 
Compared with Baseline, PureGaze achieves better performance under the same condition.
This is an advantage of our method.

ADL also achieves better performance than Baseline while it decreases the flexibility. 
ADL requires a time-consuming setup and some target samples for domain adaption.
Although recent domain adaption methods requires less samples, these user-unfriendly requirement limits the application of gaze estimation methods.
On the other hand, ADL learns specific models for each domain.
In contrast, PureGaze learns one unique model for all domains.
PureGaze cannot improve the performance in a specific domain with a large margin, for example, use a known domain-specific information to calibrate a specific model.
This is a limitation of our method.

\subsection{Visualization on Very Dark Environment.}
The visualization result of reconstruction shows our method eliminates the illumination factor.
In this section, we provide more strict cases to prove the ability of our method.

\begin{figure}[t]
	\begin{center}
		\includegraphics[width=\columnwidth]{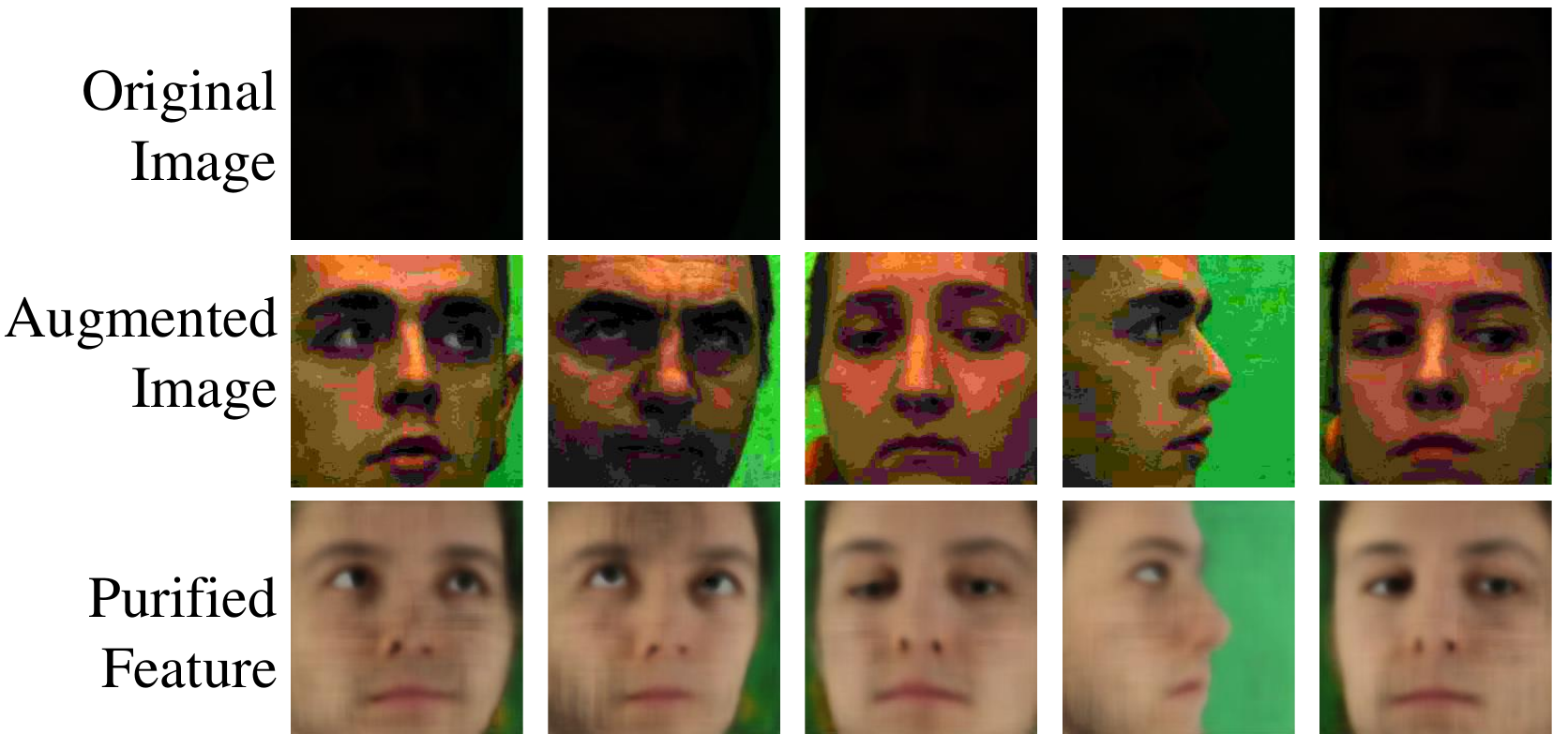}
	\end{center}
	\caption{We show the visualization result in very dark environment. The first row is the original images.  The second row is the augmented images. We manually augment these images to clearly display the image content. The third row is the reconstruction results of the purified feature. It is obvious that our method can eliminates the illumination factor in very dark environment and accurately enhance the gaze information.}
	\label{fig:dark}
\end{figure}

The original images are shown in the first row of~\Fref{fig:dark}.
Subjects in these images are invisible due to extremely dark environment.
Therefore, we manually augment these images and show them in the second row.
The reconstruction result of PureGaze is shown in the third row.
Even in very dark environment, our method still eliminates the illumination factor and captures gaze-relevant information.
In addition, compare the reconstructed images with the augmented images, it is obvious our method accurately captures the eye movement.

\subsection{Quantitative Evaluation in Illumination}
The feature reconstruction experiments show our method has the ability to remove the illumination factor from the extracted features. 
In this section, we provide the performance improvement  distribution  across  different  illumination  intensities for quantitatively analysis. 
We train the model in ETH-XGaze and test it in MPIIGaze for their rich illumination variance. 
We first cluster the images into 51 clusters according to their mean intensity. Then we remove clusters less than 7 images and compute the average accuracy.

\begin{figure}[t]
	\begin{center}
		\includegraphics[width=\columnwidth]{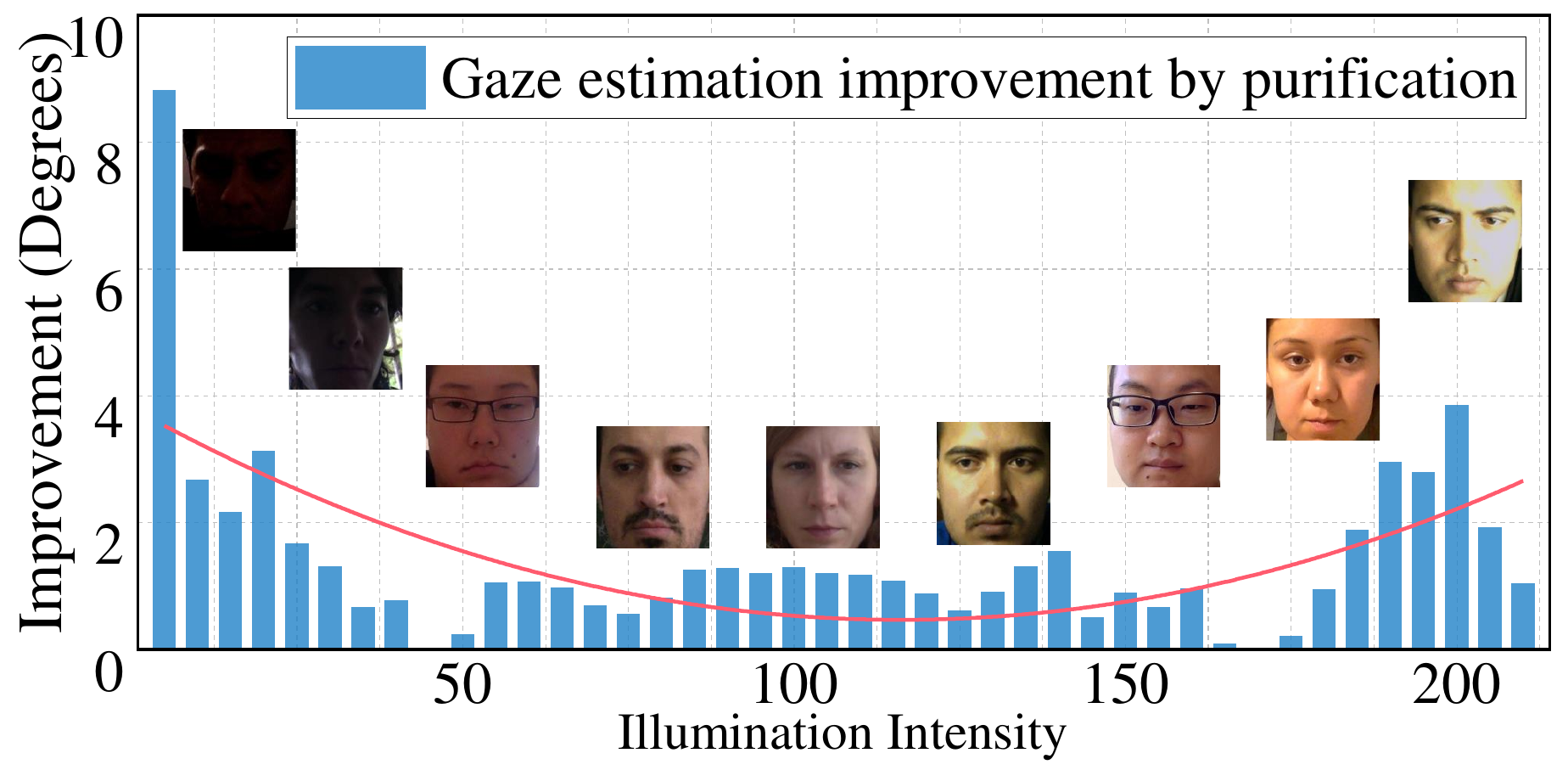}	
	\end{center}
	\caption{Quantitative evaluation in illumination. We show the performance improvement of PureGaze compared with Baseline in different illumination intensity. PureGaze largely improves the performance in the extreme illumination condition. This conclusion also matches the feature visualization result.}
	\label{fig:illumination}
\end{figure}

We illustrate the performance improvement of the PureGaze compared with the baseline in ~\Fref{fig:illumination}.
It is interesting that our method improves the performance in extreme illumination conditions.
It is because our method tries to remove the the gaze-irrelevant illumination information from the extracted feature, therefore becomes more robust than the baseline especially in extreme illumination conditions.
These results prove the advantage of the purified feature.

{\small
	\bibliographystyle{ieee_fullname}
	\bibliography{gaze}
}

\end{document}